\documentclass[10pt,twocolumn,letterpaper]{article}
\usepackage[pagenumbers]{wacv}
\usepackage{algorithm}
\usepackage{algorithmic}

\newcommand{\R}{\mathbb{R}}
\newcommand{\indic}{\mathbf{1}}
\newcommand{\cossim}{\mathrm{cos}}
\newcommand{\softmax}{\mathrm{softmax}}
\newcommand{\method}{SCCL-DT}
\DeclareMathOperator{\KL}{KL}
\DeclareMathOperator{\CE}{CE}

\usepackage{array,multirow,placeins}

\newcommand{\iTwoT}{\mathrm{I\!\to\!T}}
\newcommand{\tTwoI}{\mathrm{T\!\to\!I}}
\definecolor{wacvblue}{rgb}{0.21,0.49,0.74}
\usepackage[breaklinks,colorlinks,allcolors=wacvblue]{hyperref}
\hypersetup{pdftitle={SCCL-DT: Semantic-Conflict-Aware Contrastive Learning with Dual-Rule Teacher Distillation for Zero-Shot Chest X-Ray Diagnosis},pdfauthor={Ting Liu, Haoran Lai, HaobinChen, Ming Liu, S Kevin Zhou}}
\title{SCCL-DT: Semantic-Conflict-Aware Contrastive Learning with Dual-Rule Teacher Distillation for Zero-Shot Chest X-Ray Diagnosis}
\author{Ting Liu\textsuperscript{1,2}\quad Haoran Lai\textsuperscript{1,2}\quad HaobinChen\textsuperscript{1,2}\quad Ming Liu\textsuperscript{1,2}\quad S Kevin Zhou\textsuperscript{1,2}\\[6pt]
\parbox{0.97\textwidth}{\centering\small
\textsuperscript{1}School of Biomedical Engineering, Division of Life Sciences and Medicine,\\
University of Science and Technology of China, Hefei, Anhui, 230026, P.R. China\\[3pt]
\textsuperscript{2}Suzhou Institute for Advanced Research, University of Science and Technology of China,\\
Suzhou, Jiangsu, 215123, P.R. China}}
\date{}
\begin{document}
\maketitle

\begin{abstract}
Contrastive vision--language pre-training supports zero-shot chest X-ray diagnosis, but standard objectives treat every off-diagonal image--report pair in a batch as a negative. In radiology, recurring findings across patients mean that instance mismatch need not imply semantic incompatibility, creating false-negative supervision that we term a \emph{semantic conflict}. A training-trace audit shows that pair-observations selected by the online mask used during training are enriched for report-derived finding agreement, although this proxy analysis does not establish clinical equivalence. Fixed thresholds can become miscalibrated as similarity distributions evolve, while soft reweighting leaves residual repulsive gradients. We therefore propose \textbf{Semantic-Conflict-Aware Contrastive Learning (SCCL)}, which tracks modality-specific similarity distributions with exponential-moving-average (EMA) statistics, derives adaptive thresholds, and hard-masks flagged candidates before softmax normalization. Inclusive (OR) and consensus (AND) rules define two SCCL teachers, whose bidirectional cross-modal rankings are consolidated through standard distillation into a single deployable student, \method{}. Across five public chest X-ray benchmarks, \method{} achieves a mean zero-shot AUC of $0.850$, compared with $0.837$ for the strongest evaluated baseline in our comparison. Code will be made publicly available soon.
\end{abstract}

\section{Introduction}

Vision--language pre-training (VLP) has emerged as a promising paradigm for zero-shot chest X-ray diagnosis. By aligning radiographs with their free-text reports, VLP models learn joint representations that can be queried with disease prompts at inference without requiring task-specific annotations \citep{zhang2022convirt,huang2021gloria,tiu2022chexzero}. Most medical VLP approaches optimize a batch-wise contrastive objective \citep{oord2018cpc,radford2021clip}: within a mini-batch of $B$ image--report pairs, matched pairs are treated as positives, whereas all off-diagonal pairs are treated as negatives. This formulation inherits a key assumption from instance-level contrastive learning: every unmatched pair is a valid negative.

\begin{figure}[htbp]
\centering
\includegraphics[width=\columnwidth]{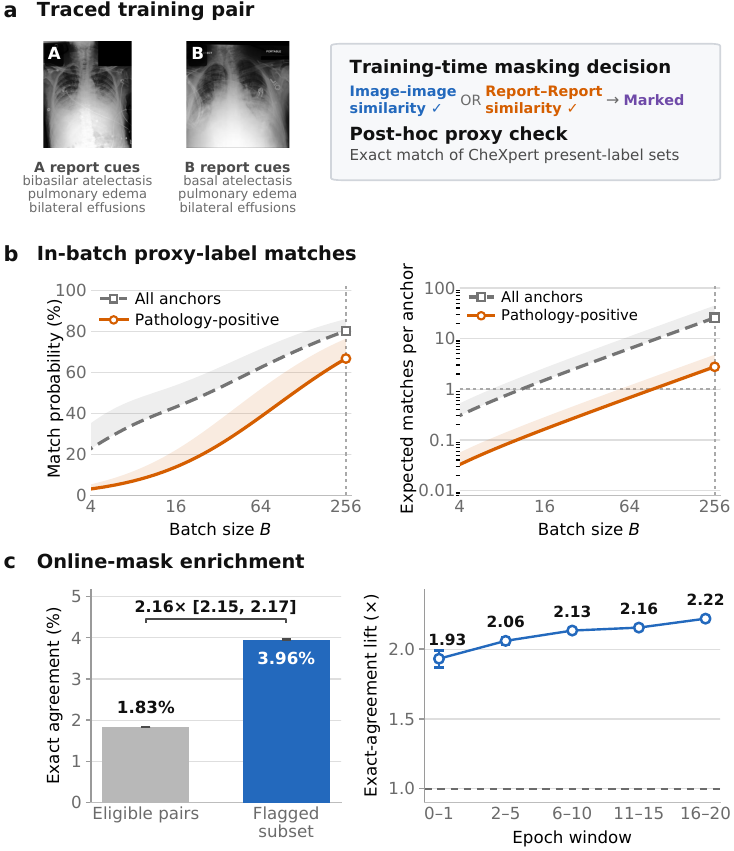}
\caption{Motivation and label-based analysis of candidate semantic conflicts.
\textbf{(a)} Online inclusive masking excludes an off-diagonal candidate from the negative set when either its image--image or report--report similarity exceeds the adaptive threshold.
\textbf{(b)} Anchor-level probability and expected count of cross-patient matches with identical present-label sets.
\textbf{(c)} Exact present-label-set agreement for flagged versus all eligible pair-observations, overall and across training windows.}
\label{fig:motivation}
\end{figure}

This assumption is particularly fragile in radiology, where reports from different patients frequently describe overlapping findings using similar or interchangeable expressions. As illustrated in Figure~\ref{fig:motivation}(a), the two batch items underlying an unmatched image--report candidate may share the same report-derived finding set. At a mini-batch size of $256$, an exact corpus-level calculation under uniform random batching estimates that $66.8\%$ of pathology-positive anchors would encounter at least one cross-patient batch mate with an identical CheXpert present-label set, with $2.78$ such matches per anchor on average (Figure~\ref{fig:motivation}(b)). A related enrichment appears in the recorded online masking decisions. Within the trace prefix ending at the best checkpoint, the inclusive rule used in the trace---which masks an off-diagonal pair when either the corresponding image--image or report--report similarity exceeds its modality-specific adaptive threshold---flags pair-observations with $3.96\%$ exact present-label-set agreement, compared with $1.83\%$ across all eligible pair-observations. Here, eligibility requires different patients and at least one positive label among the 12 pathology findings in each report; exact agreement is then evaluated over all 14 labels. The resulting $2.16\times$ enrichment remains above one throughout five consecutive training windows (Figure~\ref{fig:motivation}(c)). Although exact CheXpert present-label-set agreement does not establish clinical equivalence, its enrichment among flagged pair-observations suggests that this subset may contain a higher proportion of semantically compatible pairs. Consequently, semantically compatible pairs may still receive negative contrastive targets, giving rise to false-negative supervision. Such false negatives are known to degrade generic contrastive learning \citep{chuang2020debiased,chen2022incremental}. We refer to this mismatch between semantic compatibility and the assigned contrastive target as a \emph{semantic conflict}.

Prior work has attempted to mitigate false negatives by incorporating external clinical knowledge \citep{wang2022medclip,li2024mlip}, mining additional positive pairs from report- or study-level semantics \citep{liu2023sat,liu2024sense,zhang2026fane}, or softly reweighting suspected false negatives during optimization \citep{li2025sahn,kim2026falcon}. Despite their effectiveness, two gaps remain. \emph{Detection:} rules based on single-modality evidence or fixed, globally normalized similarity thresholds can become miscalibrated as modality-specific similarity distributions shift during representation learning. \emph{Optimization:} additional positive mining may promote partial semantic compatibility to full equivalence, whereas soft reweighting still allows suspected false negatives to exert residual repulsive gradients unless their weights become exactly zero. An effective solution must therefore identify semantic conflicts adaptively and prevent them from being pushed apart without indiscriminately relabeling them as positives.

To address both gaps, we propose \textbf{Semantic-Conflict-Aware Contrastive Learning (SCCL)}, which dynamically flags potential semantic-conflict candidates using evidence from both modalities and removes them through adaptive hard masking. For detection, SCCL maintains separate exponential-moving-average (EMA) statistics for intra-batch image--image and report--report similarities, allowing each threshold to track its modality-specific distribution as training evolves. Flagged candidates are excluded from contrastive normalization, eliminating their probability mass and repulsive gradients rather than treating them as additional positives. To account for uncertainty in the modality-specific intra-modal cues, we instantiate two rule-specific teachers: an inclusive (OR) teacher that accepts evidence from either modality to increase candidate coverage, and a consensus (AND) teacher that requires agreement between both modality-specific cues for more conservative selection. Standard distillation then consolidates the two teachers' cross-modal rankings into the single deployable student \method{}. SCCL thus adapts candidate selection to evolving similarity distributions while explicitly preventing flagged candidates from being optimized as negatives.

Our contributions are:
\begin{itemize}
    \item We propose \textbf{SCCL}, a contrastive framework that adaptively hard-masks similarity-flagged off-diagonal candidates using per-modality EMA thresholds, without adding trainable parameters to the masking module.
    \item We instantiate SCCL with inclusive and consensus rule-specific teachers and use standard distillation to consolidate their cross-modal rankings into \method{}, a single deployable student.
    \item On five public chest X-ray benchmarks (Open-I, PadChest, ChestX-ray14, CheXpert, and ChestXDet10), \method{} reaches a mean zero-shot AUC of $0.850$, the highest among the compared methods, at single-model inference cost.
\end{itemize}

\section{Related Work}

\begin{figure*}[t]
\centering
\includegraphics[width=\textwidth]{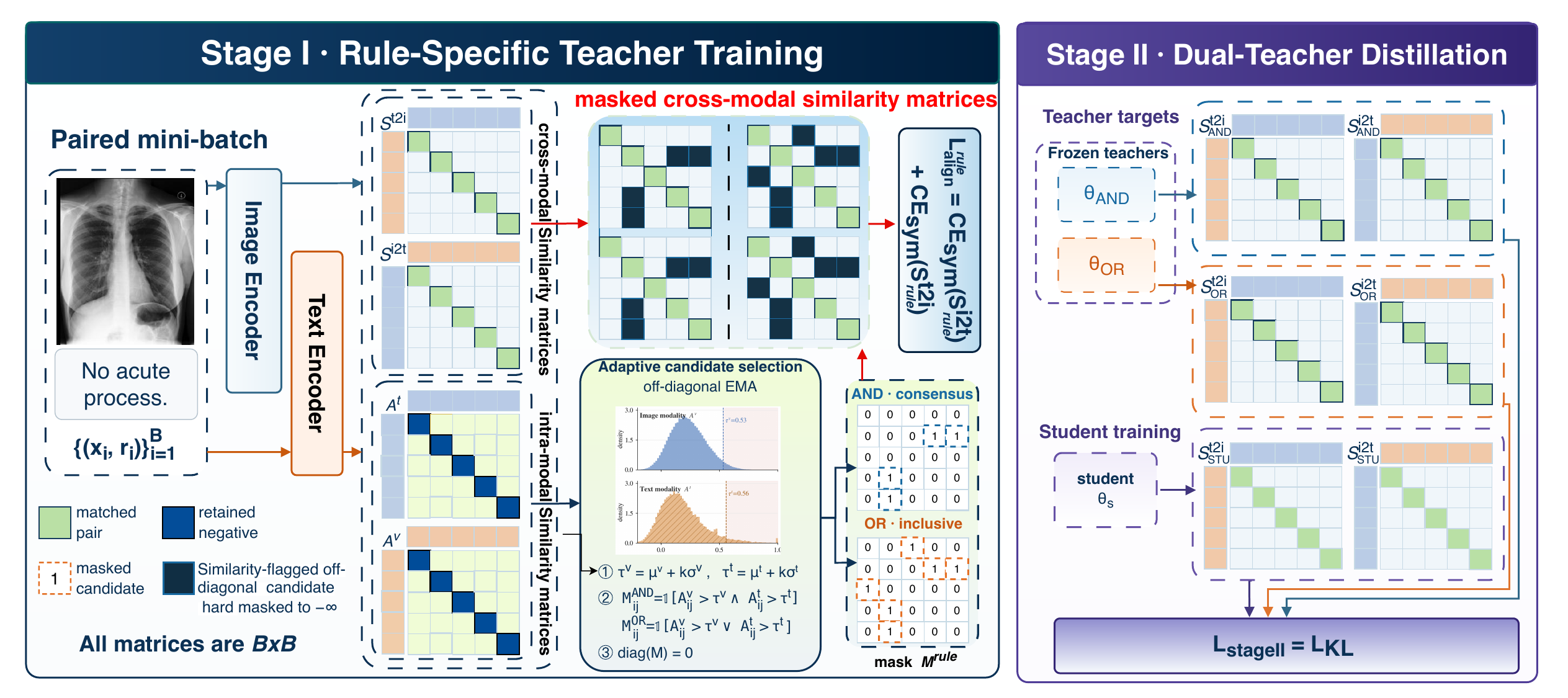}
\caption{Overview of the two-stage \method{} framework.
\textbf{Stage I} trains inclusive and consensus teachers by hard-masking similarity-flagged off-diagonal candidates selected with modality-specific EMA thresholds.
\textbf{Stage II} distills their image-to-text (i2t) and text-to-image (t2i) rankings into a single student using row- and column-wise Kullback--Leibler (KL) supervision.
Only the student is retained for inference.}
\label{fig:pipeline}
\end{figure*}

\subsection{Medical vision--language pre-training}
Contrastive alignment of chest radiographs with their reports has driven rapid progress in label-efficient and zero-shot diagnosis. ConVIRT \citep{zhang2022convirt} adapts image--text contrastive learning to radiology, while MedCLIP \citep{wang2022medclip} decouples images and texts to exploit unpaired data. GLoRIA \citep{huang2021gloria} and MGCA \citep{wang2022mgca} introduce local, multi-granularity alignment between image regions and report tokens; BioViL \citep{boecking2022biovil} strengthens the text encoder through domain-specific language modeling; and CheXzero \citep{tiu2022chexzero} demonstrates expert-level zero-shot detection. KAD \citep{zhang2023kad} incorporates a medical knowledge graph, and MRM \citep{zhou2023mrm} combines masked image and record modeling. More recent approaches replace global dot-product similarity with fine-grained cross-attention between image regions and report tokens. CARZero \citep{lai2024carzero} scores pairs through cross-attention between one modality's tokens and the other's query, while RadZero \citep{park2025radzero} extends this paradigm toward region-level, explainable alignment. Beyond these advances in broader data utilization, stronger encoders, knowledge integration, or finer-grained alignment, SCCL addresses a distinct source of supervision error: treating semantically compatible off-diagonal pairs as negatives.

\subsection{False negatives in contrastive learning}

Instance-discrimination objectives treat unmatched samples as negatives, so semantically compatible pairs may receive repulsive gradients \citep{wu2018npid,he2020moco,chen2020simclr,oord2018cpc}. Unlike valid hard negatives, false negatives should not be repelled. Prior work addresses this issue through distribution correction, hard-negative mining, progressive false-negative removal, or hardness scheduling \citep{chuang2020debiased,robinson2021hardnegatives, chen2022incremental,kim2026falcon}. 

Medical VLP methods instead exploit clinical knowledge or report semantics. MedCLIP and MLIP construct knowledge-guided targets \citep{wang2022medclip,li2024mlip}; SAT, SENSE, and FaNe use report- or study-level similarity to mine or reweight pairs \citep{liu2023sat,liu2024sense,zhang2026fane}; and SAHN separates hard from false negatives using embedding and semantic similarity \citep{li2025sahn}. SCCL differs by applying modality-specific adaptive thresholds, hard-masking selected pairs without relabeling them, and distilling its rule-specific teachers into one student.

\subsection{Knowledge distillation}
Knowledge distillation transfers predictive distributions from a teacher to a student, commonly to compress a model or retain teacher behavior in a more deployment-efficient network \citep{hinton2015distilling}. Deep mutual learning extends this idea to online peer supervision \citep{zhang2018deepmutual}, while ALBEF uses a momentum teacher to provide stable targets for noisy image--text alignment \citep{li2021albef}. These formulations differ in teacher construction but use distillation to transfer or stabilize predictive behavior. In SCCL, standard distillation consolidates two rule-specific teachers—an inclusive teacher and a consensus teacher—into a single student for efficient inference.

\section{Methodology}
\label{sec:method}

Figure~\ref{fig:pipeline} summarizes the two-stage training framework of
\method{}, from rule-specific teacher training to single-student distillation.

\subsection{Preliminaries: Batch-Contrastive Alignment}
\label{sec:method-prelim}

Let a mini-batch contain $B$ image--report pairs $\{(x_i, r_i)\}_{i=1}^{B}$, with global embeddings $v_i,t_i\in\R^{d}$. A pairwise aligner $g$ produces bidirectional score matrices $S^{\mathrm{i2t}}, S^{\mathrm{t2i}}\in\R^{B\times B}$, where $S^{\mathrm{i2t}}_{ij}$ scores image $i$ against report $j$, and $S^{\mathrm{t2i}}_{ij}$ scores report $i$ against image $j$. The diagonal contains matched pairs. Because SCCL operates on these score matrices, $g$ may be implemented by similarity, a matching head, or cross-attention; a single-matrix aligner is handled analogously.

For the bidirectional instantiation, training minimizes a symmetric cross-entropy that treats each matrix as a $B$-way classification with the diagonal as the target label $y_i=i$:
\begin{equation}
\CE_{\mathrm{sym}}(S) = \tfrac{1}{2}\big[\CE(S, y) + \CE(S^{\top}, y)\big],
\label{eq:cesym}
\end{equation}
\begin{equation}
\mathcal{L}_{\mathrm{align}}(S^{\mathrm{i2t}},S^{\mathrm{t2i}}) = \CE_{\mathrm{sym}}(S^{\mathrm{i2t}}) + \CE_{\mathrm{sym}}(S^{\mathrm{t2i}}).
\label{eq:align}
\end{equation}
Every off-diagonal entry is therefore supervised as a negative. SCCL modifies this supervision only at the score-matrix level and does not alter the internal architecture of $g$.

\subsection{Semantic-Conflict-Aware Contrastive Learning}
\label{sec:method-sccl}

\paragraph{Semantic-conflict candidates.} An off-diagonal pair becomes a semantic-conflict candidate when its negative target may conflict with partial semantic compatibility. Because pair-level clinical adjudication is unavailable, SCCL treats similarity-flagged pairs only as candidates: it removes them from softmax normalization without relabeling them as positives.

\paragraph{Per-modality similarity.} Given the batch's global embeddings, SCCL computes intra-modal cosine-similarity matrices
\begin{equation}
A^{v}_{ij}=\cossim(v_i, v_j), \qquad A^{t}_{ij}=\cossim(t_i, t_j).
\end{equation}
Let $\Omega=\{(i,j): i\neq j\}$ be the off-diagonal index set.

\paragraph{Adaptive thresholds via EMA.} At each training step, SCCL maintains modality-specific EMA estimates of the batch mean and standard deviation of the off-diagonal similarities, using momentum $m$:
\begin{align}
\mu^{v}\!\leftarrow\! m\mu^{v}\!+\!(1{-}m)\,\overline{A^{v}_{\Omega}}, \quad
\sigma^{v}\!\leftarrow\! m\sigma^{v}\!+\!(1{-}m)\,\mathrm{std}(A^{v}_{\Omega}),
\end{align}
and analogously for $\mu^{t},\sigma^{t}$. The modality-specific thresholds are
\begin{equation}
\tau^{v}=\mu^{v}+k\,\sigma^{v}, \qquad \tau^{t}=\mu^{t}+k\,\sigma^{t},
\label{eq:thresh}
\end{equation}
where $k$ determines how far into the upper tail an entry must fall. Separate EMA statistics allow the thresholds to follow each modality's evolving similarity scale.
The four EMA states are initialized from the first mini-batch. During the first $T_w$ steps, they continue to update while masking is disabled ($M=\mathbf{0}$). Statistics and masks are stop-gradient and device-local, without cross-device gathering. Validation freezes the EMA states and counters. Thresholds are not clipped; values above the cosine ceiling simply select no candidate.

\paragraph{Candidate-selection rules.} SCCL uses two candidate-selection rules: the inclusive (OR) rule accepts evidence from either modality, whereas the consensus (AND) rule requires agreement between the two modality-specific cues:
\begin{align}
M^{\mathrm{OR}}_{ij}  &= \indic\!\big[A^{v}_{ij}>\tau^{v} \ \lor\ A^{t}_{ij}>\tau^{t}\big], \label{eq:or}\\
M^{\mathrm{AND}}_{ij} &= \indic\!\big[A^{v}_{ij}>\tau^{v} \ \land\ A^{t}_{ij}>\tau^{t}\big], \label{eq:and}
\end{align}
with the diagonal forced to $0$.
Under identical thresholds, $M^{\mathrm{AND}}_{ij}\leq M^{\mathrm{OR}}_{ij}$, so $\mathcal{C}_{\mathrm{AND}}\subseteq\mathcal{C}_{\mathrm{OR}}$. OR therefore favors candidate coverage, whereas AND is more conservative. Without candidate-level ground truth, this is an expected coverage--precision trade-off rather than measured recall or precision.

\paragraph{Hard masking.} For a given teacher, let $M\in\{M^{\mathrm{OR}},M^{\mathrm{AND}}\}$ denote its rule-specific mask. Flagged candidates are removed before the softmax:
\begin{equation}
\tilde S^{q}_{ij} =
\begin{cases}
-\infty & M_{ij}=1,\\
S^{q}_{ij} & \text{otherwise},
\end{cases}
\qquad q\in\{\mathrm{i2t},\mathrm{t2i}\},\label{eq:hard-mask}
\end{equation}
where $-\infty$ is implemented as $-10^{4}$. Because $A^{v}$ and $A^{t}$ are symmetric, both rules operate on the unordered pair $\{i,j\}$ and yield $M_{ij}=M_{ji}$. Under this symmetric pair-level proxy, flagging $\{i,j\}$ provides no direction-specific evidence for retaining only one cross-swapped pairing, so we exclude both $(x_i,r_j)$ and $(x_j,r_i)$. Specifically, the former occupies $S^{\mathrm{i2t}}_{ij}$ and $S^{\mathrm{t2i}}_{ji}$, while the latter occupies $S^{\mathrm{i2t}}_{ji}$ and $S^{\mathrm{t2i}}_{ij}$. Hence a teacher applies the same rule-specific $M$ to both score matrices, removing these candidates from all row- and column-wise softmaxes. This does not mean that the two teachers share a mask: the inclusive and consensus teachers use $M^{\mathrm{OR}}$ and $M^{\mathrm{AND}}$, respectively. The training loss then replaces each $S^{q}$ by $\tilde S^{q}$ in Eq.~\eqref{eq:align}, driving the masked candidates' softmax weights and gradients to zero. The masking module adds no trainable parameters; its cost is two $B\times B$ similarity matrices and four scalar EMA updates per step. Algorithm~\ref{alg:sccl} summarizes one step.

\begin{algorithm}[t]
\caption{One training step for a rule-specific SCCL teacher.}
\label{alg:sccl}
\textbf{Input}: logits $S^{\mathrm{i2t}},S^{\mathrm{t2i}}$; embeddings $\{v_i\},\{t_i\}$; step $s$\\
\textbf{Param}: $k$, momentum $m$, warm-up $T_w$, rule $\in\{\text{OR},\text{AND}\}$\\
\textbf{Output}: masked loss $\mathcal{L}$
\begin{algorithmic}[1]
\STATE $A^{v}\!\leftarrow\!\cossim(V,V)$; \ $A^{t}\!\leftarrow\!\cossim(T,T)$
\STATE update $\mu^{v},\sigma^{v},\mu^{t},\sigma^{t}$ by EMA over off-diagonal of $A^{v},A^{t}$
\IF{$s < T_w$}
  \STATE $M \leftarrow \mathbf{0}$ \COMMENT{warm-up: no masking}
\ELSE
  \STATE $\tau^{v}\!\leftarrow\!\mu^{v}\!+\!k\sigma^{v}$; \ $\tau^{t}\!\leftarrow\!\mu^{t}\!+\!k\sigma^{t}$
  \STATE $M \leftarrow$ Eq.~\eqref{eq:or} or Eq.~\eqref{eq:and}; \ set diagonal of $M$ to $0$
\ENDIF
\STATE $\tilde S^{\mathrm{i2t}}\!\leftarrow\! S^{\mathrm{i2t}}$ masked by $M$; \ $\tilde S^{\mathrm{t2i}}\!\leftarrow\! S^{\mathrm{t2i}}$ masked by $M$
\STATE $\mathcal{L}\leftarrow \CE_{\mathrm{sym}}(\tilde S^{\mathrm{i2t}})+\CE_{\mathrm{sym}}(\tilde S^{\mathrm{t2i}})$
\STATE \textbf{return} $\mathcal{L}$
\end{algorithmic}
\end{algorithm}

\subsection{Dual-Rule Teacher Distillation}
\label{sec:method-fusion}

\paragraph{Rule-specific teachers.} Training with $M^{\mathrm{OR}}$ and $M^{\mathrm{AND}}$ yields an \emph{inclusive teacher} $\theta_{\mathrm{OR}}$ and a \emph{consensus teacher} $\theta_{\mathrm{AND}}$, respectively. We train one teacher under each rule and use both as sources for distillation.

\paragraph{Dual-rule distillation.} We train a single student $\theta_s$ to match the ensemble of the two \emph{frozen} teachers. The student is initialized independently rather than from either teacher or an average of their parameters; teacher knowledge is transferred only through logits computed on the fly for each batch, with no logits precomputed or stored. The teacher targets are the element-wise average of the two teachers' raw logits,
\begin{equation}
\bar S^{\mathrm{i2t}} = \tfrac{1}{2}\big(S^{\mathrm{i2t}}_{\mathrm{OR}}+S^{\mathrm{i2t}}_{\mathrm{AND}}\big), \quad
\bar S^{\mathrm{t2i}} = \tfrac{1}{2}\big(S^{\mathrm{t2i}}_{\mathrm{OR}}+S^{\mathrm{t2i}}_{\mathrm{AND}}\big),
\end{equation}
and the student matches them with a distillation loss at temperature $\tau_{d}$, using both row- and column-normalized distributions of both score matrices:
\begin{equation}
\begin{aligned}
\mathcal{L}_{\mathrm{KL}}
&= \frac{\tau_{d}^{2}}{4}
\sum_{\substack{q\in\{\mathrm{i2t},\mathrm{t2i}\}\\ a\in\{\mathrm{row},\mathrm{col}\}}} \\
&\quad \KL\!\Big(
\softmax_{a}\!\big(\bar S^{q}/\tau_{d}\big)
\Big\|
\softmax_{a}\!\big(S_s^{q}/\tau_{d}\big)
\Big),
\end{aligned}
\label{eq:kl}
\end{equation}
where $a$ denotes the normalization axis. The Stage-II student is trained solely with $\mathcal{L}_{\mathrm{KL}}$, so the teachers' contrastively learned bidirectional rankings provide its complete supervision. We refer to the resulting single-pass model as \method{}.

\paragraph{Zero-shot inference.} At test time, we retain only the student. For each disease $c$, we construct $p_c$ using the fixed template ``There is [disease]''. For an image $x$ and a disease prompt $p_c$, the student encodes both inputs, including the global and any aligner-specific local features, and obtains bidirectional scores $S^{\mathrm{i2t}}_{\theta_s}(x,p_c)$ and $S^{\mathrm{t2i}}_{\theta_s}(p_c,x)$. The final score for disease $c$ is
\begin{equation}
s_c(x)=\tfrac{1}{2}\!\left[S^{\mathrm{i2t}}_{\theta_s}(x,p_c)+S^{\mathrm{t2i}}_{\theta_s}(p_c,x)\right].
\end{equation}
We use this raw averaged score for multi-label evaluation, without a class-wise softmax; the teachers, EMA statistics, and candidate masks are absent at inference.

\begin{table*}[!t]
\centering
\small
\setlength{\tabcolsep}{5pt}
\begin{tabular}{lcccccc}
\toprule
Method & Open-I & PadChest & ChestX-ray14 & CheXpert & ChestXDet10 & Mean \\
\midrule
MedCLIP~\citep{wang2022medclip} & 0.551 & 0.508 & 0.564 & 0.744 & 0.571 & 0.588 \\
BiomedCLIP~\citep{zhang2023biomedclip} & 0.577 & 0.513 & 0.639 & 0.677 & 0.630 & 0.607 \\
GLoRIA~\citep{huang2021gloria} & 0.589 & 0.565 & 0.610 & 0.750 & 0.645 & 0.632 \\
BioViL~\citep{boecking2022biovil} & 0.702 & 0.655 & 0.729 & 0.789 & 0.708 & 0.717 \\
CheXzero~\citep{tiu2022chexzero} & 0.726 & 0.648 & 0.712 & 0.889 & 0.640 & 0.723 \\
MedKLIP~\citep{wu2023medklip} & 0.759 & 0.629 & 0.726 & 0.879 & 0.713 & 0.741 \\
KAD~\citep{zhang2023kad} & 0.807 & 0.750 & 0.789 & 0.905 & 0.735 & 0.797 \\
CARZero~\citep{lai2024carzero} & 0.838 & 0.810 & \underline{0.811} & \textbf{0.923} & \underline{0.796} & 0.836 \\
RadZero~\citep{park2025radzero} & \underline{0.851} & \textbf{0.841} & 0.807 & 0.903 & 0.785 & \underline{0.837} \\
\midrule
\textbf{\method{} (Ours)} & \textbf{0.853} & \underline{0.827} & \textbf{0.835} & \underline{0.922} & \textbf{0.811} & \textbf{0.850} \\
\bottomrule
\end{tabular}
\caption{Zero-shot classification performance (AUC) on five chest X-ray benchmarks.
Mean is the unweighted dataset average; higher is better.
Bold and underline denote the best and second-best results in each column, respectively.}
\label{tab:main}
\end{table*}

\section{Experiments}
\label{sec:experiments}

\subsection{Setup}

\paragraph{Implementation details.}
Our implementation builds on CARZero \citep{lai2024carzero}, which serves as the unmasked Base model. We pre-train on frontal AP/PA MIMIC-CXR images with usable reports \citep{johnson2019mimiccxr}. Available report fields are concatenated and truncated or padded to $97$ tokens. Images are resized and zero-padded to $256{\times}256$, cropped to $224{\times}224$, and augmented with random horizontal flips, affine transformations, and brightness/contrast jitter. Each run uses a ViT-B/16 image encoder initialized from M3AE \citep{dosovitskiy2021vit,chen2022m3ae} and a BioClinical BERT text encoder \citep{alsentzer2019clinicalbert}. Teachers use $k{=}2.0$, $m{=}0.99$, and $T_w{=}500$, and are optimized with Adam (learning rate $5{\times}10^{-5}$, weight decay $10^{-6}$). We halve the learning rate after five validation epochs without improvement, early-stop after ten, and select the lowest-loss checkpoint. The student is trained for six epochs with Adam at $10^{-5}$, using the same weight decay and scheduler and $\tau_d{=}2.0$. All training runs are conducted on NVIDIA A800 GPUs. The masking operator itself maintains only four scalar EMA states and adds no trainable weights. The complete two-stage protocol nevertheless trains two rule-specific teachers and one student under supervision from the two frozen teachers; only the student is retained for inference.

\paragraph{Zero-shot evaluation.} We evaluate on five public chest X-ray datasets without any fine-tuning: \textbf{Open-I} \citep{demnerfushman2016openi}, \textbf{PadChest} \citep{bustos2020padchest} (manually annotated subset), \textbf{ChestX-ray14} \citep{wang2017chestxray8}, \textbf{CheXpert} \citep{irvin2019chexpert}, and \textbf{ChestXDet10} \citep{liu2020chestxdet10}. For multi-label zero-shot classification, AUC is the primary metric; the PadChest ablations additionally report MCC, F1, and accuracy (ACC). For zero-shot grounding on ChestXDet10, we report the unweighted mean Pointing Game accuracy across ten findings. The $14{\times}14$ cross-attention map is expanded to $224{\times}224$ by patch-wise replication, and a trial is counted as a hit when a maximum-attention region overlaps an annotated lesion. Each map is independently min--max normalized for visualization.

\paragraph{Baselines.} We compare against MedCLIP \citep{wang2022medclip}, BiomedCLIP \citep{zhang2023biomedclip}, GLoRIA \citep{huang2021gloria}, BioViL \citep{boecking2022biovil}, CheXzero \citep{tiu2022chexzero}, MedKLIP \citep{wu2023medklip}, KAD \citep{zhang2023kad}, CARZero \citep{lai2024carzero}, and RadZero \citep{park2025radzero}. For the zero-shot classification comparison in Table~\ref{tab:main}, we re-evaluate all baselines using their publicly released models on the same five official test sets.

\paragraph{Candidate-set analysis.}
Across the recorded training trace, the actual OR mask enriches report-derived exact label agreement from \(1.83\%\) among eligible pairs to \(3.96\%\) among flagged pairs (\(2.16\times\)). Descriptively, the lift is above one in each of five epoch windows and is also observed for Image-only candidates; the same qualitative pattern is observed under partial-overlap measures and an alternative labeler. These analyses evaluate whether the online mask preferentially selects label-overlapping pairs; they do not measure the recall of clinically adjudicated false negatives. Full definitions, counts, and sensitivity analyses are provided in the supplementary material.

\subsection{Main Results}

\paragraph{Zero-shot classification.}
Table~\ref{tab:main} compares \method{} with published zero-shot chest X-ray classification methods across five benchmarks. The final distilled student achieves the highest mean AUC of $0.850$, exceeding RadZero ($0.837$) and CARZero ($0.836$) by $0.013$ and $0.014$, respectively. Under this unified classification evaluation protocol, \method{} achieves the highest AUC on Open-I, ChestX-ray14, and ChestXDet10, while ranking second on PadChest and CheXpert; on CheXpert, it trails the best result by only $0.001$. The consistent performance across these heterogeneous datasets provides evidence of cross-dataset robustness, while inference retains a single model.

CARZero and RadZero provide the closest radiology-specific comparisons. \method{} outperforms each on four of the five datasets: relative to CARZero, the gains are $+0.015$ on Open-I, $+0.017$ on PadChest, $+0.024$ on ChestX-ray14, and $+0.015$ on ChestXDet10, whereas the two methods are nearly tied on CheXpert ($0.922$ vs.\ $0.923$). Relative to RadZero, \method{} performs better on all datasets except PadChest and improves the mean AUC by $0.013$. Accordingly, Table~\ref{tab:main} provides a same-test-set, public-checkpoint classification comparison under a common dataset and label scope. The controlled ablations below further isolate the effect of semantic-conflict masking within a shared implementation.

\begin{figure*}[!t]
\centering
\includegraphics[width=0.98\textwidth]{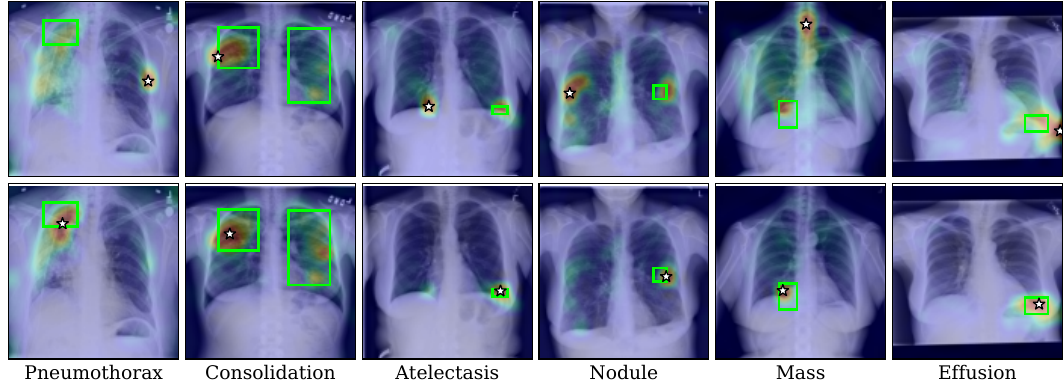}
\caption{Qualitative zero-shot grounding on six ChestXDet10 cases. Rows show the unmasked Base model and \method{}; green boxes are ground-truth lesions, white stars are Pointing Game maxima, and warmer colors indicate larger cross-attention weights. Each map is normalized independently.}
\label{fig:attention}
\end{figure*}

\paragraph{Zero-shot grounding.} Table~\ref{tab:grounding} reports both mean and per-finding Pointing Game accuracy on ChestXDet10. \method{} achieves the highest mean accuracy of $0.583$, exceeding the CARZero ($0.543$) and RadZero at $224$ pixels ($0.537$) values by $4.0$ and $4.6$ percentage points, respectively. Relative to each of the tabulated CARZero and RadZero rows, \method{} improves on seven of ten findings, matches one, and is lower on two. Across all included methods, it obtains the highest or tied-highest accuracy on six findings. Figure~\ref{fig:attention} complements these per-finding results with six qualitative Base-versus-\method{} comparisons. Since a Pointing Game hit requires a maximally attended region to overlap an annotated lesion, these results provide evidence of stronger zero-shot spatial correspondence between cross-attention and radiographic abnormalities, rather than detection or segmentation performance.

\begin{table*}[!t]
\centering
\scriptsize
\setlength{\tabcolsep}{3.2pt}
\resizebox{\textwidth}{!}{%
\begin{tabular}{lccccccccccc}
\toprule
Method & Mean & ATE & CALC & CONS & EFF & EMPH & FIB & FX & MASS & NOD & PTX \\
\midrule
GLoRIA~\citep{huang2021gloria}
& 0.368 & 0.479 & 0.053 & 0.737 & 0.528 & 0.667 & 0.366 & 0.013 & 0.533 & 0.156 & 0.143 \\
MedKLIP~\citep{wu2023medklip}
& 0.481 & 0.625 & 0.132 & \underline{0.837} & 0.675 & 0.734 & 0.305 & \underline{0.224} & \underline{0.733} & 0.312 & 0.229 \\
KAD~\citep{zhang2023kad}
& 0.391 & \underline{0.646} & 0.132 & 0.699 & 0.618 & 0.644 & 0.244 & 0.199 & 0.267 & 0.316 & 0.143 \\
CARZero~\citep{lai2024carzero}
& \underline{0.543} & 0.604 & \underline{0.184} & 0.824 & 0.782 & \textbf{0.846} & \textbf{0.561} & 0.184 & 0.700 & 0.286 & \underline{0.457} \\
RadZero (224~px)~\citep{park2025radzero}
& 0.537 & 0.604 & \textbf{0.211} & 0.806 & \textbf{0.813} & 0.795 & 0.451 & 0.197 & \textbf{0.767} & \underline{0.325} & 0.400 \\
\midrule
\method{} (Ours)
& \textbf{0.583} & \textbf{0.667} & \underline{0.184} & \textbf{0.855}
& \underline{0.802} & \underline{0.821} & \underline{0.524} & \textbf{0.263}
& \textbf{0.767} & \textbf{0.351} & \textbf{0.600} \\
\bottomrule
\end{tabular}%
}
\caption{Zero-shot grounding on ChestXDet10, measured by per-finding and unweighted mean Pointing Game (PG) accuracy. ATE, CALC, CONS, EFF, EMPH, FIB, FX, MASS, NOD, and PTX denote Atelectasis, Calcification, Consolidation, Effusion, Emphysema, Fibrosis, Fracture, Mass, Nodule, and Pneumothorax, respectively. Bold and underline denote the best and second-best results in each column.}
\label{tab:grounding}
\end{table*}

\begin{table}[!b]
\centering
\small
\setlength{\tabcolsep}{2pt}
\begin{tabular}{c|ccc|cccc|c}
\toprule
 & \multicolumn{3}{c|}{Components} & \multicolumn{4}{c|}{PadChest} & Cross-data \\
\cmidrule(lr){2-4}\cmidrule(lr){5-8}\cmidrule(l){9-9}
\# & OR & AND & KD & AUC $\uparrow$ & MCC $\uparrow$ & F1 $\uparrow$ & ACC $\uparrow$ & mAUC $\uparrow$ \\
\midrule
(a) &              &              &              & 0.810 & 0.156 & 0.143 & 0.991 & 0.836 \\
(b) & $\checkmark$ &              &              & \underline{0.826} & \textbf{0.179} & \textbf{0.165} & 0.991 & 0.840 \\
(c) &              & $\checkmark$ &              & 0.825 & 0.173 & 0.159 & 0.991 & \underline{0.846} \\
(d) & $\checkmark$ &              & $\checkmark$ & 0.823 & \underline{0.178} & 0.162 & 0.991 & \underline{0.846} \\
(e) &              & $\checkmark$ & $\checkmark$ & 0.822 & 0.169 & 0.155 & 0.991 & 0.844 \\
(f) & $\checkmark$ & $\checkmark$ & $\checkmark$ & \textbf{0.827} & \textbf{0.179} & \underline{0.164} & 0.991 & \textbf{0.850} \\
\bottomrule
\end{tabular}
\caption{Rule-specific teacher and knowledge-distillation analysis. OR/AND denote the teacher(s), KD denotes student distillation, and the empty setting is the unmasked Base. Cross-data mAUC averages five benchmarks; bold and underline mark best and second-best; ACC ties after rounding.}
\label{tab:abl-rule}
\end{table}

\subsection{Ablation and Diagnostic Studies}

We report controlled PadChest studies under a unified evaluation protocol across AUC, MCC, F1, and accuracy (ACC). The OR/AND/KD, threshold, and similarity-cue studies characterize prespecified SCCL configurations. Two-tier, count-matched Random, and Soft weighting instead serve as diagnostic controls rather than SCCL components: they examine whether a single compromise rule can replace the separate inclusive and consensus teachers, whether gains depend on structured candidate selection beyond mask quantity alone, and whether hard exclusion is preferable to soft attenuation, respectively. These results are not used for hyperparameter tuning or default-configuration selection. The five-dataset mean AUC provides a cross-dataset consistency check. We retain ACC as a critical classification metric; because it remains near $0.991$ across configurations, most between-configuration differences appear in AUC, MCC, and F1. Unless otherwise specified, the controlled studies use a prespecified default comprising an adaptive threshold with $k{=}2$, the inclusive rule, hard masking, and both similarity cues; Table~\ref{tab:abl-rule} separately compares the two rule-specific teachers and the single- and dual-teacher students.

\paragraph{Teacher and student configurations.} Both SCCL teachers improve over the unmasked Base model in PadChest AUC, MCC, and F1 (Table~\ref{tab:abl-rule}). Among the single-teacher students, distillation from the inclusive teacher obtains a higher cross-dataset mAUC than distillation from the consensus teacher ($0.846$ vs. $0.844$; rows (d) and (e)). Using both rule-specific teachers raises mAUC to $0.850$ (row (f)), while achieving the highest PadChest AUC and tying the highest MCC. The inclusive teacher remains marginally higher in F1. Distillation therefore serves model consolidation rather than the primary methodological contribution.

\begin{table}[t]
\centering
\small
\setlength{\tabcolsep}{3pt}
\begin{tabular}{lcccc}
\toprule
Design choice & AUC $\uparrow$ & MCC $\uparrow$ & F1 $\uparrow$ & ACC $\uparrow$ \\
\midrule
\multicolumn{5}{l}{\textbf{A. Threshold sensitivity}}\\
\quad Fixed $\tau{=}0.85$          & 0.807 & 0.147 & 0.130 & 0.991 \\
\quad Fixed $\tau{=}0.90$          & 0.812 & 0.155 & 0.136 & 0.991 \\
\quad Fixed $\tau{=}0.95$          & 0.814 & 0.156 & 0.138 & 0.991 \\
\quad Adaptive $k{=}1.5$           & \underline{0.820} & 0.157 & 0.139 & 0.991 \\
\quad Adaptive $k{=}2$ (def.)      & \textbf{0.826} & \textbf{0.179} & \textbf{0.165} & 0.991 \\
\quad Adaptive $k{=}2.5$           & 0.819 & \underline{0.159} & \underline{0.141} & 0.991 \\
\midrule
\multicolumn{5}{l}{\textbf{B. Diagnostic comparisons}}\\
\multicolumn{5}{l}{\emph{Candidate-selection logic}}\\
\quad Inclusive (def.)             & \textbf{0.826} & \textbf{0.179} & \textbf{0.165} & 0.991 \\
\quad Consensus                    & \underline{0.825} & \underline{0.173} & \underline{0.159} & 0.991 \\
\quad Two-tier                     & 0.818 & 0.162 & 0.146 & 0.991\\
\quad Random (count-matched)        & 0.817 & 0.156 & 0.139 & 0.991 \\
\addlinespace
\multicolumn{5}{l}{\emph{Logit intervention}}\\
\quad Hard masking (def.)           & \textbf{0.826} & \textbf{0.179} & \textbf{0.165} & 0.991 \\
\quad Soft weighting               & \underline{0.820} & \underline{0.161} & \underline{0.139} & 0.991 \\
\bottomrule
\end{tabular}
\caption{Controlled PadChest analyses: (A) threshold sensitivity; (B) candidate-selection and logit-intervention controls. ``def.'' marks the prespecified default; bold and underline mark best and second-best within each block; ACC ties after rounding.}
\label{tab:abl-robust}
\end{table}

\begin{table}[t]
\centering
\small
\setlength{\tabcolsep}{4pt}
\begin{tabular}{lcccc}
\toprule
Similarity cue & AUC $\uparrow$ & MCC $\uparrow$ & F1 $\uparrow$ & ACC $\uparrow$ \\
\midrule
Image only     & 0.814 & \underline{0.158} & \underline{0.141} & 0.991 \\
Text only      & \underline{0.816} & 0.157 & 0.138 & 0.991 \\
Both (default) & \textbf{0.826} & \textbf{0.179} & \textbf{0.165} & 0.991 \\
\bottomrule
\end{tabular}
\caption{PadChest similarity-cue ablation under the default inclusive setting. Bold and underline mark the best and second-best AUC, MCC, and F1; ACC ties at $0.991$.}
\label{tab:abl-cue}
\end{table}

\paragraph{Controlled design analyses.}
Table~\ref{tab:abl-robust} combines threshold sensitivity with diagnostic comparisons around the prespecified default configuration: adaptive $k{=}2$, inclusive selection, hard masking, and joint cues.

\begin{figure}[!t]
\centering
\includegraphics[width=0.90\columnwidth]{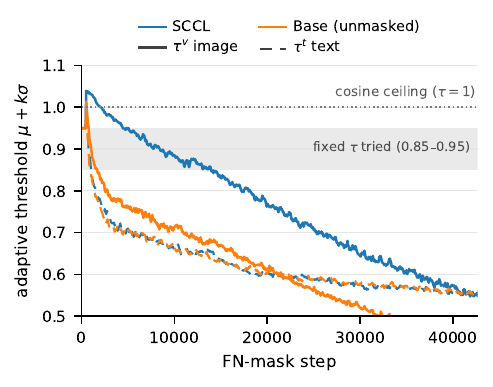}
\caption{Adaptive image (solid) and text (dashed) thresholds ($k{=}2$) for SCCL and the unmasked Base. Base values are logged but not applied; shading marks the fixed range $0.85$--$0.95$ and the dotted line the cosine ceiling.}
\label{fig:thresh}
\end{figure}

\noindent\emph{Threshold rule.}
Panel A compares the implemented fixed- and adaptive-threshold variants. Under the implemented fixed branch, thresholds \(\tau\leq0.80\) are omitted because early OR-mask saturation removes nearly all off-diagonal negatives and yields a degenerate near-zero loss. The adaptive rules use the device-local EMA statistics in Eq.~\eqref{eq:thresh}, frozen during validation. Adaptive $k{=}2$ achieves an AUC/MCC/F1 of $0.826/0.179/0.165$, compared with $0.814/0.156/0.138$ for the strongest non-degenerate fixed threshold ($\tau{=}0.95$). Figure~\ref{fig:thresh} shows that the adaptive thresholds track the evolving similarity scale and converge to approximately $0.55$ late in training. Among the tested adaptive settings, the prespecified $k{=}2$ gives the highest AUC, MCC, and F1 point estimates on PadChest.

\noindent\emph{Candidate-selection rule.}
Panel B compares diagnostic controls with the prespecified references. The inclusive and consensus rules are defined in Sec.~\ref{sec:method-sccl}. Two-tier combines stringent single-modality evidence with moderate agreement across modalities, while Random uniformly samples off-diagonal pairs to match the total per-batch inclusive mask budget. Inclusive obtains the highest AUC/MCC/F1 point estimates of $0.826/0.179/0.165$. Consensus nearly preserves AUC ($0.825$) but reduces MCC/F1 to $0.173/0.159$, whereas Two-tier performs worse across all three metrics. Together with Table~\ref{tab:abl-rule}, this comparison supports retaining the inclusive and consensus rules as separate teachers rather than collapsing them into a single heuristic. Count-matched random masking obtains $0.817/0.156/0.139$, supporting the importance of SCCL's structured candidate selection beyond mask quantity alone.

\noindent\emph{Masking operation and modality cues.}
Within Panel B, Soft weighting serves as a diagnostic control that attenuates likely conflicts rather than removing them. Hard masking improves AUC/MCC/F1 from $0.820/0.161/0.139$ with soft weighting to $0.826/0.179/0.165$, supporting complete removal of the associated probability mass under the tested setting. Combining image and text cues also outperforms either image-only or text-only selection across all three metrics (Table~\ref{tab:abl-cue}), indicating that the two modalities provide complementary evidence. Full definitions of the three diagnostic controls are provided in the supplement.

\section{Conclusion}

Standard contrastive pre-training treats all off-diagonal image--report pairs as negatives, which can introduce semantic conflicts when reports share findings. We proposed \textbf{Semantic-Conflict-Aware Contrastive Learning (SCCL)}, which uses modality-specific adaptive thresholds to identify candidate conflicts and removes them from contrastive normalization without relabeling them as positives. Distillation consolidates the resulting rule-specific supervision into a single deployable model. The online mask is enriched for report-derived finding agreement, providing proxy evidence for the selection mechanism rather than pair-level clinical adjudication. Across five public benchmarks, \method{} achieves the highest mean zero-shot AUC among the evaluated methods ($0.850$ vs.\ $0.837$) and the highest mean ChestXDet10 Pointing Game accuracy ($0.583$). Together, these results support adaptive conflict removal as an effective alternative to indiscriminate negative supervision under the evaluated settings.

Future work should clinically adjudicate candidate pairs, study calibration across batch sizes and domains, and extend adaptive detection to local region--token alignment and other medical imaging modalities.

{\small
\bibliographystyle{ieeenat_fullname}
\bibliography{main}
}
\clearpage
\appendix
\raggedbottom
\setcounter{dbltopnumber}{4}
\renewcommand{\dbltopfraction}{0.9}
\setcounter{table}{0}
\renewcommand{\thetable}{S\arabic{table}}
\renewcommand{\theHtable}{S\arabic{table}}
\setcounter{equation}{0}
\renewcommand{\theequation}{S\arabic{equation}}
\renewcommand{\theHequation}{S\arabic{equation}}
\section*{Supplementary Material}
\begingroup
\sloppy
This supplement documents the data and evaluation scope, the Stage-II
student-distillation objective, the Pointing Game evaluation protocol, implementation-level
definitions of the Two-tier, count-matched Random, and Soft-weighting
ablations, the candidate-set analysis, and the failure mode observed for low
fixed thresholds. The descriptions below distinguish the proposed SCCL
components from dataset and evaluation choices that remain fixed across all
model variants.

\section{Data and Evaluation Scope}
\label{sec:supp-data}

\subsection{Pre-training data}

We pre-train on frontal AP/PA MIMIC-CXR images with usable report content,
using the archived training and validation manifests and report filters
associated with the Base implementation. The same manifests, filters, and
data-loading configuration are used for the Base model, both rule-specific
teachers, and the distilled student. SCCL changes the treatment of candidate
negatives during pre-training; it does not redefine the dataset partition or
report selection.

\subsection{Downstream evaluation scope}

We use the same five zero-shot evaluation sets and target-label mappings as in
the main paper: Open-I \citep{demnerfushman2016openi}; the
radiologist-annotated subset of PadChest \citep{bustos2020padchest}; the
official NIH test split of ChestX-ray14 \citep{wang2017chestxray8}; the
CheXpert test set of 500 studies from 500 patients, evaluated on the five
competition observations \citep{irvin2019chexpert}; and the official
ChestXDet10 test split with box annotations for ten findings
\citep{liu2020chestxdet10}. ChestXDet10 is used for both zero-shot
classification and grounding; the other four datasets are used for
classification. All compared models are evaluated without fine-tuning using their publicly
released checkpoints and inference code on the same benchmark subsets,
target-label mappings, and metrics.

\section{Stage-II Student Distillation Objective}
\label{sec:supp-stage-two-distillation}

Let $S_{\mathrm{OR}}^{q}$ and $S_{\mathrm{AND}}^{q}$ denote the raw logits from
the frozen inclusive and consensus teachers for
$q\in\{\iTwoT,\tTwoI\}$. Their element-wise average defines the teacher target,
\begin{equation}
\bar S^{q}=\frac{1}{2}\left(S_{\mathrm{OR}}^{q}+S_{\mathrm{AND}}^{q}\right).
\label{eq:supp-teacher-target}
\end{equation}
For the student logits $S_s^{q}$, we match both row- and column-normalized
distributions at temperature $\tau_d$:
\begin{equation}
\begin{aligned}
\mathcal L_{\mathrm{KL}}
&=\frac{\tau_d^2}{4}
\sum_{\substack{q\in\{\iTwoT,\tTwoI\}\\
a\in\{\mathrm{row},\mathrm{col}\}}} \\
&\quad
\operatorname{KL}\!\left(
\operatorname{softmax}_{a}(\bar S^{q}/\tau_d)
\,\middle\|\,
\operatorname{softmax}_{a}(S_s^{q}/\tau_d)
\right).
\end{aligned}
\label{eq:supp-stage-two-kl}
\end{equation}
The Stage-II objective is pure distillation,
\begin{equation}
\mathcal L_{\mathrm{Stage\ II}}=\mathcal L_{\mathrm{KL}},
\qquad \tau_d=2.0,
\label{eq:supp-stage-two-objective}
\end{equation}
with no additional student loss. The two teachers are frozen throughout this
stage, and only the student is retained for inference.

\section{Grounding Evaluation Details}
\label{sec:supp-pg}

We evaluate zero-shot grounding on the ChestXDet10 test set using the Pointing
Game (PG). For each image--finding trial, the cross-attention map produced for
the corresponding disease prompt serves as the grounding prediction. A trial
is counted as a hit when a maximally attended region overlaps the annotated
lesion region. We compute PG independently for each of the ten findings and
report their unweighted mean. The complete mean and per-finding comparison is
reported in the main paper.

\section{Diagnostic-Control Definitions and Stage-I Training Objective}
\label{sec:supp-ablation-definitions}

This section precisely defines the Two-tier, count-matched Random, and Soft
weighting controls used to interpret the main results. They are explanatory
comparisons rather than removable components of SCCL: Two-tier examines a
single compromise rule for the inclusive/consensus decision logic, Random
controls for mask quantity, and Soft weighting contrasts attenuation with hard
exclusion. Each changes only the within-batch logits passed to the
contrastive cross-entropy during stage-one pre-training; none changes an
encoder, the fusion module, or the inference path. The definitions follow the
active MLP stage-one implementation and the configuration used by each run.

For a batch of $B$ image--report pairs, let $v_i$ and $t_i$ be
the global image and text embeddings. We first normalize them and construct
the two \emph{within-modality}, rather than cross-modal, similarity matrices:
\begin{equation}
\bar v_i=
\frac{v_i}{\max(\lVert v_i\rVert_2,10^{-12})},
\qquad
\bar t_i=
\frac{t_i}{\max(\lVert t_i\rVert_2,10^{-12})},
\end{equation}
\begin{equation}
A^{v}_{ij}=\bar v_i^\top\bar v_j,\qquad
A^{t}_{ij}=\bar t_i^\top\bar t_j.
\label{eq:supp-within-modality-similarity}
\end{equation}
Let $\Omega=\{(i,j):i\neq j\}$ and $N=|\Omega|=B(B-1)$. Threshold-based
branches compute, separately for each modality, the batch mean and sample
standard deviation over all \emph{ordered} off-diagonal cells:
\begin{equation}
\begin{aligned}
\mu_s^\ell&=\frac{1}{N}\sum_{(i,j)\in\Omega}A^\ell_{ij},\\
\sigma_s^\ell&=
\sqrt{\frac{1}{N-1}\sum_{(i,j)\in\Omega}(A^\ell_{ij}-\mu_s^\ell)^2},
\quad \ell\in\{v,t\}.
\end{aligned}
\label{eq:supp-batch-statistics}
\end{equation}
Thus both $(i,j)$ and $(j,i)$ enter the statistics. The code uses PyTorch
\texttt{std()}, i.e., a sample standard deviation. It maintains separate EMAs
of the mean and of the standard deviation itself, not an EMA of variance.
The first update initializes each state with the current batch statistic;
subsequent updates are
\begin{equation}
\begin{aligned}
\widehat\mu_s^\ell&=m\widehat\mu_{s-1}^\ell+(1-m)\mu_s^\ell,\\
\widehat\sigma_s^\ell&=m\widehat\sigma_{s-1}^\ell+(1-m)\sigma_s^\ell,
\quad m=0.99.
\end{aligned}
\label{eq:supp-ema}
\end{equation}
When state updates are enabled, the current batch enters
Eq.~\eqref{eq:supp-ema} before its threshold is computed. Validation calls with
frozen state reuse the latest training EMA. Variant-specific call-counting
semantics are given below.

The fusion module produces two raw score matrices
$S^{\iTwoT},S^{\tTwoI}\in\R^{B\times B}$. For any matrix $X$, with the
batch diagonal $y_i=i$ as the positive pairing, define
\begin{equation}
\operatorname{CE}_{\mathrm{row}}(X)=
-\frac{1}{B}\sum_{i=1}^{B}
\log\frac{\exp(X_{ii})}{\sum_{j=1}^{B}\exp(X_{ij})},
\end{equation}
\begin{equation}
\mathcal D(X)=\frac{1}{2}\left[
\operatorname{CE}_{\mathrm{row}}(X)+
\operatorname{CE}_{\mathrm{row}}(X^\top)\right].
\label{eq:supp-dqncos}
\end{equation}
All three variants optimize
\begin{equation}
\mathcal L=\lambda_{\mathrm{ce}}\left[
\mathcal D(\widetilde S^{\iTwoT})+
\mathcal D(\widetilde S^{\tTwoI})\right],
\qquad \lambda_{\mathrm{ce}}=1.
\label{eq:supp-total-loss}
\end{equation}
The two directions are summed, not averaged; within each direction the row and
column CE terms are averaged by Eq.~\eqref{eq:supp-dqncos}. Both directions
reuse the same fusion module. The active implementation disables local/global
auxiliary losses, so each variant affects Eq.~\eqref{eq:supp-total-loss} only
through $\widetilde S$.

\subsection{Two-tier: modality-asymmetric two-level rule}

Two-tier uses high and low thresholds with different image- and text-side
multipliers:
\begin{align}
\tau^{v}_{H,s}&=\widehat\mu_s^{v}+3.0\,\widehat\sigma_s^{v},
&
\tau^{t}_{H,s}&=\widehat\mu_s^{t}+2.0\,\widehat\sigma_s^{t},
\label{eq:supp-two-tier-high}\\
\tau^{v}_{L,s}&=\widehat\mu_s^{v}+1.5\,\widehat\sigma_s^{v},
&
\tau^{t}_{L,s}&=\widehat\mu_s^{t}+1.0\,\widehat\sigma_s^{t}.
\label{eq:supp-two-tier-low}
\end{align}
For an off-diagonal pair, the high tier accepts strong evidence in either
modality, whereas the low tier requires agreement between modalities:
\begin{align}
M^{\mathrm{strong}}_{s,ij}
&=\indic[A^{v}_{ij}>\tau^{v}_{H,s}]
\lor\indic[A^{t}_{ij}>\tau^{t}_{H,s}],
\label{eq:supp-two-tier-strong}\\
M^{\mathrm{weak}}_{s,ij}
&=\indic[A^{v}_{ij}>\tau^{v}_{L,s}]
\land\indic[A^{t}_{ij}>\tau^{t}_{L,s}].
\label{eq:supp-two-tier-weak}
\end{align}
The final mask is
\begin{equation}
M^{\mathrm{TT}}_{s,ij}=
\begin{cases}
M^{\mathrm{strong}}_{s,ij}\lor M^{\mathrm{weak}}_{s,ij},&i\neq j,\\
0,&i=j.
\end{cases}
\label{eq:supp-two-tier-mask}
\end{equation}
All comparisons are strict; equality to a threshold does not trigger. The
thresholds are not clipped to $[-1,1]$, and there is no cap on candidate count.
Because the matrices in Eq.~\eqref{eq:supp-within-modality-similarity} are
symmetric, $M^{\mathrm{TT}}$ is symmetric. The implementation nevertheless
uses its transpose explicitly in the t2i branch to preserve the score-axis
semantics.

The actual parameters are $m=0.99$ and a warm-up of
$T_w=500$ \emph{training masker calls}. The first $500$ training
calls update the EMA but return an all-zero mask. The $501$st call first updates
the EMA and then applies Eq.~\eqref{eq:supp-two-tier-mask}. The constructor
parameter \texttt{init\_threshold=0.95} is displayed in warm-up logs only and
does not enter a decision. Validation freezes the EMA, step, and trigger-rate
state; after warm-up, validation loss still uses the current training EMA to
form a mask.

Two-tier replaces logits rather than subtracting a finite offset:
\begin{align}
\widetilde S^{\iTwoT}_{ij}=
\begin{cases}-10^4,&M^{\mathrm{TT}}_{s,ij}=1,\\
S^{\iTwoT}_{ij},&\text{otherwise},\end{cases}
\nonumber\\
\widetilde S^{\tTwoI}_{ij}=
\begin{cases}-10^4,&M^{\mathrm{TT}}_{s,ji}=1,\\
S^{\tTwoI}_{ij},&\text{otherwise}.\end{cases}
\label{eq:supp-two-tier-injection}
\end{align}
The diagonal is always preserved. Mask generation runs under
\texttt{no\_grad}, so no gradient passes from the similarity decision branch to
the global embeddings.

\subsection{Count-matched Random}

At training call $s$, the reference inclusive run records the total number of
unordered pairs selected in the upper triangle,
\begin{equation}
K_s=\sum_{1\leq i<j\leq B_s}M^{\mathrm{ref}}_{s,ij}.
\label{eq:supp-random-target-count}
\end{equation}
Random replays this budget in the same training-call order. For
$\mathcal U_s=\{\{i,j\}:1\leq i<j\leq B_s\}$, it samples exactly
$K_s$ pairs uniformly without replacement:
\begin{equation}
\mathcal R_s\sim\operatorname{Uniform}
\{\mathcal S\subseteq\mathcal U_s:|\mathcal S|=K_s\}.
\label{eq:supp-random-sampling}
\end{equation}
The sampled pairs are mirrored to form a symmetric mask with a zero diagonal
and are injected using the hard replacement in
Eq.~\eqref{eq:supp-two-tier-injection}. Sampling uses a private CPU random
generator with base seed $1$. The reference and Random runs otherwise use the
same training configuration and batch schedule. This control exactly matches
the total mask budget at every call, but not per-row or per-column counts or
row degree; it therefore compares SCCL's structured candidate selection with
an unstructured count-matched mask.

\subsection{Soft weighting: continuous Noisy-OR retention}

Soft weighting uses
\begin{equation}
\begin{aligned}
\tau_s^\ell&=\widehat\mu_s^\ell+2.0\,\widehat\sigma_s^\ell,
&\ell&\in\{v,t\},\\
\delta_s^\ell&=\max(\widehat\sigma_s^\ell,10^{-6}).
\end{aligned}
\label{eq:supp-soft-threshold}
\end{equation}
and assigns each modality a continuous false-negative probability
\begin{equation}
p^\ell_{s,ij}=\operatorname{sigmoid}\left(
\gamma\,\frac{A^\ell_{ij}-\tau_s^\ell}{\delta_s^\ell}\right),\qquad \gamma=4.0.
\label{eq:supp-soft-probability}
\end{equation}
The quantity applied to logits is a \emph{retention weight}, not the
false-negative probability:
\begin{equation}
W_{s,ij}=(1-p^{v}_{s,ij})(1-p^{t}_{s,ij}),\quad i\neq j,
\qquad W_{s,ii}=1.
\label{eq:supp-soft-keep-weight}
\end{equation}
Equation~\eqref{eq:supp-soft-keep-weight} is a De Morgan relaxation of hard OR:
a high false-negative probability in either modality reduces retention. The
configuration field \texttt{T} multiplies the standardized margin. We
therefore denote this factor by $\gamma$: mathematically, it is a slope or
inverse temperature, and larger $\gamma$ produces a sharper transition closer
to hard masking. The experiment fixes $\gamma=4.0$ with no schedule.

The weight is neither multiplied into a sample loss after CE nor multiplied
directly with a raw logit. It is added in log space after the shared fusion
module creates the $B\times B$ logits and before
Eq.~\eqref{eq:supp-dqncos}:
\begin{align}
\overline W_{s,ij}&=\max(W_{s,ij},10^{-30}),
&b_{s,ij}&=\log\overline W_{s,ij},
\label{eq:supp-soft-log-bias}\\
\widetilde S^{\iTwoT}&=S^{\iTwoT}+b_s,
&\widetilde S^{\tTwoI}&=S^{\tTwoI}+b_s^\top.
\label{eq:supp-soft-injection}
\end{align}
Although $W_s$ is theoretically symmetric, the t2i implementation still takes
its transpose explicitly. Since
\begin{equation}
\exp(S_{ij}+\log\overline W_{ij})
=\overline W_{ij}\exp(S_{ij}),
\end{equation}
the operation exactly multiplies each negative's exponentiated contribution to
the softmax denominator by the clamped $\overline W_{ij}$ while preserving the
positive through $W_{ii}=1$. A row-wise CE term is therefore
\begin{equation}
-\frac{1}{B}\sum_i\log
\frac{\exp(S_{ii})}
{\exp(S_{ii})+\sum_{j\neq i}\overline W_{ij}\exp(S_{ij})}.
\label{eq:supp-soft-ce}
\end{equation}

The actual settings are $k=2.0$, $m=0.99$,
$T_w=500$, and $\gamma=4.0$. The first $500$ training
\texttt{get\_mask} calls update the EMA but return an all-one weight matrix;
the $501$st training call first enables Eq.~\eqref{eq:supp-soft-keep-weight}.
\texttt{init\_threshold=0.95} remains log-only. Weights are computed under
\texttt{no\_grad}, so gradients pass only through the corrected fusion logits.
During validation, the EMA state and warm-up counter are frozen, and the latest
training EMA is reused without advancing either state.

\subsection{Unified execution summary}

The three variants follow the same high-level sequence:
\begin{enumerate}
\item Compute the normalized within-image and within-text matrices in
Eq.~\eqref{eq:supp-within-modality-similarity}.
\item Compute mean/std on all ordered $i\neq j$ cells and update the
modality-specific EMAs according to Eq.~\eqref{eq:supp-ema}. For all three
variants, only training calls update the EMA and advance the warm-up counter;
sanity and validation calls reuse the frozen training state.
\item During warm-up, use $M=0$ for Two-tier, $W=1$ for Soft, and read
$K_s=0$ and use $M=0$ for Random.
\item After warm-up, form $M$ using
Eqs.~\eqref{eq:supp-two-tier-high}--\eqref{eq:supp-two-tier-mask}, sample
exactly $K_s$ Random pairs using
Eq.~\eqref{eq:supp-random-sampling}, or compute
$W$ using Eqs.~\eqref{eq:supp-soft-probability}--\eqref{eq:supp-soft-keep-weight}.
\item Inject the mask/weight: Two-tier and Random replace the selected i2t and
transposed t2i cells by $-10^4$; Soft adds
$\log(\max(W,10^{-30}))$ and its transpose.
\item Compute Eq.~\eqref{eq:supp-total-loss}, averaging row/column CE within
each direction and summing the two directions.
\end{enumerate}

\begin{table*}[t]
\centering
\small
\setlength{\tabcolsep}{4pt}
\begin{tabular}{p{0.16\textwidth}p{0.45\textwidth}p{0.31\textwidth}}
\toprule
Variant & Decision/sampling parameters & Logit intervention \\
\midrule
Two-tier
& $k_H^{v}=3.0$, $k_H^{t}=2.0$;
$k_L^{v}=1.5$, $k_L^{t}=1.0$; $m=0.99$;
$T_w=500$ training-call warm-up
& Symmetric mask; i2t uses $M$, t2i uses $M^\top$; selected cells replaced by
$-10^4$ \\
\addlinespace
Count-matched Random
& OR source: $k=2.0$, $m=0.99$; $T_w=500$ training-call warm-up;
full upper-triangular pool;
exact per-call $K_s$; global seed $=1$, mask base seed $=1$
& Same hard replacement; matches only total unordered-pair count, not
row/column degree \\
\addlinespace
Soft weighting
& $k=2.0$, $m=0.99$; $T_w=500$ training-call warm-up; $\gamma=4.0$;
std floor $10^{-6}$;
pre-log clamp $10^{-30}$
& i2t adds $\log\overline W$, t2i its transpose, between fusion logits and CE \\
\bottomrule
\end{tabular}
\caption{Implementation semantics of the three diagnostic controls.}
\label{tab:supp-variant-summary}
\end{table*}

All three implementations unconditionally retain positive diagonal pairs, and
mask/weight computation itself is excluded from gradient propagation.

\section{Training-Time Candidate-Set Analysis}
\label{sec:supp-candidate-analysis}

This section separates two questions represented in Fig.~1 in the main paper. First, a
model-free corpus calculation asks how often uniform random batches contain
cross-patient samples with matching report-derived finding sets. Second, an
actual-trace analysis asks whether the inclusive OR masks used by the training
loss are enriched for report-derived agreement. The latter follows the online
decisions of the evolving model and EMA state; no final-checkpoint detector is
reconstructed for the primary analysis.

\subsection{Corpus-level collision analysis}

The filtered pre-training corpus contains
$N_{\mathrm{corpus}}=227{,}237$ image--report items. CheXpert labels are mapped
to Boolean present sets under L2/U-Zero: only a raw value of $1$ is present,
whereas $0$, $-1$, and missing values are not. A pathology-positive anchor has
at least one of the 12 findings remaining after excluding \emph{No Finding}
and \emph{Support Devices}; this gives $134{,}937$ pathology-positive anchors.

For anchor $i$, let $m_i$ be the number of other corpus items from a different
patient with the same complete 14-label present set. Conditional on placing
$i$ in a uniformly sampled batch of size $B$, the number of matching batch
mates is hypergeometric,
\begin{equation}
X_i(B)\sim\operatorname{Hypergeom}
(N_{\mathrm{corpus}}-1,m_i,B-1).
\end{equation}
The collision probability and expected count are therefore
\begin{align}
\pi_i(B)
&=1-\frac{\binom{N_{\mathrm{corpus}}-1-m_i}{B-1}}
{\binom{N_{\mathrm{corpus}}-1}{B-1}},
\label{eq:supp-collision-probability}\\
e_i(B)&=(B-1)\frac{m_i}{N_{\mathrm{corpus}}-1}.
\label{eq:supp-collision-expectation}
\end{align}
Figure~1(b) in the main paper averages Eqs.~\eqref{eq:supp-collision-probability} and
\eqref{eq:supp-collision-expectation} with equal weight over the prespecified
anchor population. Table~\ref{tab:supp-anchor-collisions-actual} gives the
training-schedule batch sizes.

\begin{table*}[t]
\centering
\small
\setlength{\tabcolsep}{5pt}
\begin{tabular}{rcccc}
\toprule
&\multicolumn{2}{c}{All anchors}&
\multicolumn{2}{c}{Pathology-positive anchors}\\
$B$ & $\widehat P(\geq1)$ & $\widehat E[X]$
& $\widehat P(\geq1)$ & $\widehat E[X]$ \\
\midrule
$64$  & $62.46\%$ & $6.37$  & $38.47\%$ & $0.69$ \\
$128$ & $72.05\%$ & $12.83$ & $53.41\%$ & $1.38$ \\
$256$ & $80.21\%$ & $25.77$ & $66.78\%$ & $2.78$ \\
\bottomrule
\end{tabular}
\caption{Exact cross-patient anchor-level collision statistics used in
Fig.~1(b) in the main paper.}
\label{tab:supp-anchor-collisions-actual}
\end{table*}

At $B=256$, \emph{No Finding} anchors account for $91.3\%$ of the expected
matches in the all-anchor calculation. The pathology-positive result
($66.78\%$ and $2.78$ matches) therefore provides the cleaner main-text
summary. As sensitivity checks, ignoring \emph{No Finding} and \emph{Support
Devices} only in the matching criterion raises the $B=256$ collision
probabilities to $86.03\%$ for all anchors and $76.47\%$ for
pathology-positive anchors; removing the cross-patient restriction changes
each exact probability by less than $0.05$ percentage point.

\subsection{Recorded online masks and analysis endpoint}

The trace stores the ordered sample identities, epoch, batch index, global and
mask steps, batch size, EMA statistics, thresholds, and the exact image-high,
text-high, and applied masks for each training batch. Symmetric masks are
represented by their strict upper triangle. For every analyzed batch, the
stored applied mask equals the union of the stored image-high and text-high
masks.

The lowest validation loss occurs at
\texttt{epoch=20-step=27504.ckpt}. Its durable trace prefix ends at
\texttt{mask\_step=27504}; after excluding the 500-step warm-up, the primary
endpoint is therefore \texttt{mask\_step=500,...,27504}. This range contains
$27{,}005$ batches from logged epochs 0--20: $6{,}600$ batches of size 64,
$7{,}100$ of size 128, and $13{,}305$ of size 256. Their complete strict-upper-
triangle spaces contain $505{,}289{,}600$ pair-observations.

The primary CheXpert pool requires both samples to contain at least one of the
12 pathology findings and to come from different patients. This gives
$N=178{,}150{,}296$ pair-observations. Exact agreement compares the complete
14-label present set after the pool is selected. NegBio is evaluated both on
its analogous labeler-defined pool and on the fixed CheXpert pool.

\subsection{Estimands and uncertainty}

For eligible pair-observation $r$, let $M_r$ denote the stored applied-mask
decision and $A_r$ exact present-set agreement. We accumulate
\begin{align}
N&=\sum_r 1, & F&=\sum_r M_r,\\
G&=\sum_r A_r, & FG&=\sum_r M_rA_r.
\label{eq:supp-trace-counts}
\end{align}
The flagged agreement rate, eligible-pair baseline, lift, trigger rate, and
agreement capture rate are
\begin{align}
p_{\mathrm{flag}}&=FG/F,
&p_{\mathrm{base}}&=G/N,
&L&=p_{\mathrm{flag}}/p_{\mathrm{base}},\\
T&=F/N,
&R&=FG/G.
\label{eq:supp-trace-estimands}
\end{align}
The baseline contains all eligible observations in the same batches and pool,
including flagged observations; it is not a separately sampled control set.
As a within-batch composition check, we also compute
\begin{equation}
\mathbb E_0[FG]=\sum_b F_bG_b/N_b,
\qquad
\mathrm{O/E}=\frac{\sum_b FG_b}{\mathbb E_0[FG]}.
\end{equation}

Confidence intervals are estimated by an epoch-stratified batch-level
bootstrap over the recorded training trace.

\subsection{Exact-agreement results}

Table~\ref{tab:supp-primary-actual} reports the complete primary counts. The
online OR mask flags $6{,}613{,}862$ of $178{,}150{,}296$ CheXpert-pool
observations; $261{,}620$ of the flagged observations agree exactly. The
resulting $3.9556\%$ flagged rate, $1.8340\%$ baseline, and
$2.1568\times$ lift give an $8.0071\%$ capture rate. The batch-conditioned
expectation is $121{,}400.07$ agreements, compared with $261{,}620$ observed,
for O/E $=2.1550$ ($95\%$ CI, $2.1455$--$2.1646$).

\begin{table*}[t]
\centering
\scriptsize
\setlength{\tabcolsep}{3.2pt}
\begin{tabular}{lrrrrrrrr}
\toprule
Pool / agreement & $N$ & $F$ & $G$ & $FG$ & Flagged & Baseline
& Lift ($95\%$ CI) & Capture \\
\midrule
CheXpert / CheXpert
& 178,150,296 & 6,613,862 & 3,267,352 & 261,620
& 3.9556\% & 1.8340\% & 2.1568 (2.1471--2.1664) & 8.0071\% \\
NegBio / NegBio
& 170,566,834 & 6,323,181 & 3,306,661 & 255,078
& 4.0340\% & 1.9386\% & 2.0809 (2.0715--2.0902) & 7.7141\% \\
CheXpert / NegBio (fixed pool)
& 178,150,296 & 6,613,862 & 3,304,696 & 255,801
& 3.8676\% & 1.8550\% & 2.0850 (2.0755--2.0943) & 7.7405\% \\
\bottomrule
\end{tabular}
\caption{Best-checkpoint-aligned enrichment using the masks actually applied
during training. Counts are repeated pair-observations along the trace.}
\label{tab:supp-primary-actual}
\end{table*}

\subsection{Temporal persistence}

The right panel of Fig.~1(c) in the main paper groups the best-checkpoint-aligned trace into five
consecutive epoch windows. Table~\ref{tab:supp-time-windows} records the batch
schedule and the complete strict-upper-triangle observation budget for each
window. Lift ranges from $1.93$ to $2.22$, and every window-level bootstrap
interval remains above one. Because batch size changes with epoch, the points
establish persistence rather than a monotone learning trend.

\begin{table*}[t]
\centering
\small
\setlength{\tabcolsep}{4pt}
\begin{tabular}{lrrr}
\toprule
Epoch window & Batch size & Batches & All $i<j$ observations / lift \\
\midrule
0--1   & 64  & 6,600 & 13,305,600 / $1.93\times$ \\
2--5   & 128 & 7,100 & 57,708,800 / $2.06\times$ \\
6--10  & 256 & 4,435 & 144,758,400 / $2.13\times$ \\
11--15 & 256 & 4,435 & 144,758,400 / $2.16\times$ \\
16--20 & 256 & 4,435 & 144,758,400 / $2.22\times$ \\
\bottomrule
\end{tabular}
\caption{Five-window temporal analysis displayed in Fig.~1(c) in the main paper. All
window-level $95\%$ CI lower bounds exceed one.}
\label{tab:supp-time-windows}
\end{table*}

\subsection{Image, text, OR, and AND rules}

Table~\ref{tab:supp-four-rules} evaluates four decisions on the same fixed
CheXpert pool. Image and Text are the corresponding high-threshold masks, OR
is their inclusive union actually used by the analyzed teacher, and AND is
their intersection. Image- and text-triggered sets are not disjoint; their
intersection is the AND row. The OR result is therefore not explained only by
the text rule.

\begin{table*}[t]
\centering
\small
\setlength{\tabcolsep}{4pt}
\begin{tabular}{lrrrrrr}
\toprule
Rule & $F$ & $FG$ & Agreement & Baseline & Lift & Trigger / capture \\
\midrule
Image & 1,348,802 & 95,719 & 7.0966\% & 1.8340\% & 3.8694
& 0.7571\% / 2.9296\% \\
Text & 5,355,304 & 175,084 & 3.2694\% & 1.8340\% & 1.7826
& 3.0061\% / 5.3586\% \\
OR & 6,613,862 & 261,620 & 3.9556\% & 1.8340\% & 2.1568
& 3.7125\% / 8.0071\% \\
AND & 90,244 & 9,183 & 10.1757\% & 1.8340\% & 5.5483
& 0.0507\% / 0.2811\% \\
\bottomrule
\end{tabular}
\caption{Four-rule exact CheXpert agreement on the fixed primary pool.
Trigger is $F/N$ and capture is $FG/G$.}
\label{tab:supp-four-rules}
\end{table*}

The mutually exclusive source partition gives image-only, text-only, and both
shares of $19.03\%$, $79.61\%$, and $1.36\%$ of OR flags, respectively. Their
exact-agreement lifts are $3.7490$ ($95\%$ CI, $3.7190$--$3.7788$), $1.7181$
($1.7088$--$1.7273$), and $5.5483$ ($5.4333$--$5.6626$). This is an
associative source decomposition rather than a causal modality ablation.

\subsection{Partial finding overlap}

Exact equality is intentionally strict. On the same eligible observations,
the online OR set also shows greater partial overlap: $57.4\%$ of flagged
observations share at least one present finding, compared with $45.5\%$ of all
eligible observations, while mean Jaccard similarity is $0.252$ versus $0.181$
(Table~\ref{tab:supp-partial-overlap}). Thus the exact-agreement result is not
solely a consequence of the equality threshold.

\begin{table}[t]
\centering
\small
\setlength{\tabcolsep}{4pt}
\begin{tabular}{lrrrr}
\toprule
Metric & Flagged & Eligible & Difference & Ratio \\
\midrule
Shared any finding & 57.4\% & 45.5\% & $+11.9$ pp & $1.26\times$ \\
Mean Jaccard & 0.252 & 0.181 & $+0.071$ & $1.39\times$ \\
\bottomrule
\end{tabular}
\caption{Partial-overlap checks on the fixed CheXpert primary pool.}
\label{tab:supp-partial-overlap}
\end{table}

\subsection{Alternative pools and full-trace sensitivity}

Enrichment remains above one under every alternative CheXpert-defined pool
(Table~\ref{tab:supp-alt-pools-actual}). Removing the different-patient
condition changes the pathology-positive estimate only from $2.1568$ to
$2.1597$.

\begin{table*}[t]
\centering
\small
\setlength{\tabcolsep}{4pt}
\begin{tabular}{lrrrrr}
\toprule
Pair pool & $N$ & $F$ & Flagged & Baseline & Lift ($95\%$ CI) \\
\midrule
All eligible $i<j$ pairs
& 505,289,600 & 27,476,966 & 35.6397\% & 10.1071\%
& 3.5262 (3.5196--3.5329) \\
Exclude pairs where both are No Finding
& 444,327,034 & 17,348,162 & 2.0469\% & 1.0037\%
& 2.0394 (2.0313--2.0474) \\
Both samples have No Finding $\ne1$
& 215,233,750 & 7,968,418 & 4.4564\% & 2.0720\%
& 2.1508 (2.1422--2.1593) \\
Both pathology-positive; no patient exclusion
& 178,166,296 & 6,619,910 & 3.9619\% & 1.8344\%
& 2.1597 (2.1501--2.1693) \\
\bottomrule
\end{tabular}
\caption{Sensitivity to the CheXpert-defined pair pool.}
\label{tab:supp-alt-pools-actual}
\end{table*}

Extending the descriptive calculation to the full post-warm-up trace through
\texttt{mask\_step=36374} yields $280{,}227{,}481$ CheXpert-pool observations
and a $2.1774\times$ lift ($3.9940\%$ vs. $1.8343\%$). The analogous
NegBio-defined analysis contains $268{,}294{,}947$ observations and yields
$2.1013\times$ ($4.0745\%$ vs. $1.9390\%$). These are sensitivity estimates;
the best-checkpoint-aligned prefix remains the primary endpoint.

\subsection{Interpretation}

The actual online mask is persistently enriched for report-derived exact and
partial finding agreement, including under image-triggered decisions and a
second automatic labeler. This evidence concerns association within one
recorded training trajectory. The absolute exact-agreement rate is $3.9556\%$,
the OR capture rate is $8.0071\%$, and no independent manual clinical
adjudication has been completed. The analysis therefore does not establish
pair-level clinical equivalence or a causal downstream benefit of masking.

\section{Fixed-Threshold Mask Saturation}
\label{sec:supp-fixed-threshold}

The implemented fixed-threshold branch returns an all-zero mask for the first
$500$ masker calls and then applies one shared image/text cutoff:
\begin{equation}
M^{\mathrm{fix}}_{s,ij}=\indic[i\neq j]\,
\indic[A^{v}_{ij}>\tau\ \lor\ A^{t}_{ij}>\tau].
\label{eq:supp-fixed-mask}
\end{equation}
The warm-up counter follows masker calls rather than optimizer steps. Selected
i2t logits are replaced by $-10^4$, t2i uses the transposed mask, and positive
diagonal pairs are always retained.

In the executed runs with $\tau\in\{0.50,0.55,0.80\}$, activation of the
fixed OR mask removed nearly all off-diagonal negatives and drove the
contrastive loss close to zero. At the audited $\tau=0.80$ representation,
the image branch selected every off-diagonal pair, so any lower cutoff is no
less permissive. Runs with $\tau\in\{0.85,0.90,0.95\}$ remained
non-degenerate. We therefore omit the saturated low-threshold settings from
downstream evaluation and report $\tau=0.95$ as the best-performing
non-degenerate fixed setting.

This behavior occurs when an absolute cutoff enters the bulk of the current
similarity distribution. In contrast, SCCL's modality-specific thresholds
$\tau_s^\ell=\widehat\mu_s^\ell+k\widehat\sigma_s^\ell$ track the evolving image
and text similarity scales, reducing the risk of such saturation.

\endgroup
\end{document}